\pdfoutput=1

\documentclass[11pt]{article}

\usepackage[final]{acl}

\usepackage{times}
\usepackage{latexsym}
\usepackage{amssymb}
\usepackage{float}

\usepackage[T1]{fontenc}

\usepackage[utf8]{inputenc}

\usepackage{microtype}

\usepackage{inconsolata}
\usepackage{booktabs}
\usepackage{soul}
\usepackage{color}

\usepackage{multirow}
\usepackage{multicol}
\usepackage{enumitem,kantlipsum}
\usepackage[normalem]{ulem}
\usepackage{color,colortbl}
\usepackage{todonotes} 
\usepackage{soul}
\usepackage{threeparttable, makecell}

\usepackage{makecell}
\usepackage{xcolor}

\usepackage{tcolorbox}

\definecolor{lightblue}{rgb}{.50,.95,1}
\definecolor{tri}{rgb}{.25,.88,.82}
\definecolor{lilac}{rgb}{0.85,0.64,0.85}

\title{Large Language Models for Propaganda Span Annotation}

\author{
Maram Hasanain,  Fatema Ahmad, Firoj Alam \\
Qatar Computing Research Institute, HBKU, Qatar\\
\{mhasanain,fakter,fialam\}@hbku.edu.qa
}

\begin{document}
\maketitle

\begin{abstract}
The use of propagandistic techniques in online content has increased in recent years aiming to manipulate online audiences. Fine-grained propaganda detection and extraction of textual spans where propaganda techniques are used, are essential for more informed content consumption. Automatic systems targeting the task over lower resourced languages are limited, usually obstructed by lack of large scale training datasets. Our study investigates whether Large Language Models (LLMs), such as GPT-4, can effectively extract propagandistic spans. We further study the potential of employing the model to collect more cost-effective annotations. Finally, we examine the effectiveness of labels provided by GPT-4 in training smaller language models for the task. The experiments are performed over a large-scale in-house manually annotated dataset.
The results suggest that providing more annotation context to GPT-4 within prompts improves its performance compared to human annotators. Moreover, when serving as an expert annotator (consolidator), the model provides labels that have higher agreement with expert annotators, and lead to specialized models that achieve state-of-the-art over an unseen Arabic testing set. Finally, our work is the \textit{first} to show the potential of utilizing LLMs to develop annotated datasets for propagandistic spans detection task 
prompting it with annotations from human annotators with limited expertise. All scripts and annotations will be shared with
 the community.\footnote{\url{https://github.com/MaramHasanain/llm_prop_annot}} 
\end{abstract}

\section{Introduction}
\label{sec:intro}


Malicious actors are actively exploiting online platforms to disseminate misleading content for political, social, and economic agendas \citep{perrin2015social,alam-etal-2022-survey,ijcai2022p781}.  
 The objective of using propaganda is to generate distorted and often misleading information, which can result in heightened polarization on specific issues and division among communities.  
  Hence, it is important to automatically detect and debunk propagandistic content. The majority of relevant research has focused on either binary or multiclass and multilabel classification scenarios of the task~\cite{BARRONCEDENO20191849,rashkin-EtAl:2017:EMNLP2017,piskorski2023semeval}. More recently, interest has shifted to finer-grained propaganda detection at the text span level, which is a multilabel sequence tagging task, where more than one propaganda technique can be used within the same text span~\cite{EMNLP19DaSanMartino,DaSanMartinoSemeval20task11,propaganda-detection:WANLP2022-overview,przybyla2023long,araieval:arabicnlp2024-overview}. Such fine-grained analysis is necessary for system explainability and improved digital media literacy among news readers. The task in its nature is complex~\cite{da2020survey}, and the complexity is magnified by the large number of propaganda techniques that might be present (18~\citep{EMNLP19DaSanMartino} \textit{vs.} 23~\citep{piskorski2023semeval} techniques for example).
The subjective nature of the task also results in added challenges. 

LLMs showed remarkable capabilities on versatile downstream NLP tasks, and on a plethora of languages, including Arabic~\cite{bang2023multitask,ahuja2023mega,abdelali2023benchmarking,liang2022holistic}. However, the utility of LLMs in span-level propaganda detection and categorization remains under-explored. Therefore, we aim to leverage LLMs selecting the highly effective, GPT-4 \cite{openai2023gpt4}, for the task. Moreover, LLMs have shown to be effective aids in creating annotated datasets to train or evaluate other models in a variety of tasks~\citep{alizadeh2023opensource}. Since there are many propaganda techniques to label and a need to create large and diverse datasets to train specialized models, LLMs might benefit the process of developing new datasets for propaganda span detection.
Recruiting humans to carry such large-scale annotations has been a very tedious and costly procedure. Our study also aims to investigate whether we could use a LLM, such as GPT-4, to reduce human annotation cost and effort by either reducing the number of annotators, or hiring annotators with less expertise. Finally, to further understand the value of automatic propaganda labeling with LLMs, we employ labels generated by the model under different setups to train specialized language models for the task.

Specifically, we study the following research questions:
\emph{(i)} Is GPT-4 capable of annotating propagandistic spans effectively?
\emph{(ii)} Can GPT-4 serve both as a general and as an expert annotator of propaganda spans?\footnote{For this task, the manual annotation process followed generally has two phases: \emph{(i)} annotation done by three \textit{general} annotators, who are less experienced but trained annotators \emph{(ii)} annotations reviewed and disagreements resolved by two expert annotators, referred to as consolidators.}  \emph{(iii)} Which propaganda techniques can GPT-4 annotate best?
\emph{(iv)} Can we effectively train specialized models for the task using GPT-4's annotations?
Our study makes the following contributions:
\begin{itemize}
    \item We explore the use of GPT-4 as an annotator for detecting and labeling spans with propagandistic techniques, which is the \textit{first attempt} at such a task. Results reveal the great potential of the model to replace more expert annotators for some propaganda techniques, including those that are highly prevalent in the experimental dataset, such as ``Loaded Language''.  
    We also provide an in-depth analysis of the model performance at different annotation stages, for more informed adoption of such annotation approach.
    \item We show that when serving as a consolidator, GPT-4 provides labels that can be effectively used to train a specialized model for the task, achieving state-of-the-art performance on a recently released Arabic dataset from the ArAIEval shared task~\cite{araieval:arabicnlp2024-overview}. When testing that specialized model on the testing subset from our in-house dataset, it degraded performance by only 13\% compared to training the model with the gold labels from the training subset.
    \item We are releasing all scripts, and annotations from human annotators and GPT-4 to benefit the community.\footnote{\url{https://github.com/MaramHasanain/llm_prop_annot}}
\end{itemize}

\section{Related Work}
\label{sec:related_work}

\paragraph{Propaganda Detection.}
Relevant research has employed diverse methods to identify propagandistic text, ranging from analyzing content based on writing style and readability features in articles \cite{rashkin-EtAl:2017:EMNLP2017,BARRONCEDENO20191849} to using transformer based models for classification at the binary, multiclass, and multilabel settings \cite{SemEval2021-6-Dimitrov}. 
Recent efforts stress the importance of fine-grained identification of specific propagandistic techniques~\cite{DaSanMartinoSemeval20task11}. \citet{EMNLP19DaSanMartino} identified 18 distinct techniques and created a dataset by manually annotating English news articles based on them. 
 Next, they designed a multi-granular deep neural network that extracts propagandistic spans from sentences with a limited F$_1$=22.58, showing how complex the task is. 
\citet{piskorski2023semeval} extended the 18 techniques into 23 and introduced a dataset in multiple languages. 
With these efforts, fine-grained propaganda detection in general, and over Arabic content and other lower-resourced languages specifically, still requires further exploration. Existing Arabic datasets are limited in size and number of targeted techniques~\cite{propaganda-detection:WANLP2022-overview,hasanain-etal-2023-araieval}. 

\paragraph{LLMs as Annotators.}
Constructing high-quality annotated datasets, essential for model training and evaluation, usually requires manual annotation by humans~\cite{khurana2023natural}.  
There has been efforts in utilizing LLMs for data annotation to overcome the challenges of human annotations, which include bias, time-overhead, and cost \cite{ding-etal-2023-gpt,alizadeh2023opensource,thomas2023large}. 
\citet{sprenkamp2023large} investigated the effectiveness of LLMs in annotating propaganda by utilizing five variations of GPT-3 and GPT-4. 
They tackled the task as a multi-label classification problem, using the SemEval-2020 Task 11 dataset.
Their findings indicate that GPT-4 achieves results comparable to the current state of the art. Our work is closely related to theirs, however, they approached the problem as a multi-label text classification task of 14 techniques at the article level. In contrast, we focus on fine-grained propaganda detection at the span level including both multilabel and sequence tagging tasks, covering 23 techniques, which is much more challenging. 

\section{Dataset}
\label{sec:dataset}
Existing Arabic datasets for span-level propaganda detection either lack text span-level annotations (e.g., ArAIEval 2023 shared task dataset \cite{hasanain-etal-2023-araieval}), or cover a more limited set of propaganda techniques (e.g., \cite{propaganda-detection:WANLP2022-overview}).~\footnote{A large-scale Arabic dataset was released in parallel to this work as part of the ArAIEval 2024 shared task \cite{araieval:arabicnlp2024-overview}} Furthermore, to explore the potential of using LLMs as propagandistic spans annotators, a comprehensive dataset with complete human annotations is required as a gold standard. 

For this study, an in-house developed dataset is utilized, referred to as \textbf{\emph{ArPro}} across this work. It includes a total of \textbf{8,000} annotated paragraphs among which, 63\% contain at least one propagandistic span. The paragraphs were selected from 2.8$K$ news articles, with approximately 10$K$ sentences, and around 277$K$ words. It covers 14 different topics, with `news' and `politics' accounting for over 50\% of paragraphs. We split the dataset in a stratified manner~\cite{sechidis2011stratification}, allocating 75\%, 8.5\%, and 16.5\% for training, development, and testing, respectively. We briefly discuss the dataset development process. A complete detail of that process is provided in our recent work~\cite{hasanain2024can}.  

The dataset construction started from a large in-house collection of Arabic news articles sourced from over 300 Arabic news media, and including over 600K articles. We sample a set of 2.8K articles following a stratified sampling approach over the news media. Thus, we ensure a versatile set, featuring a variety of writing styles and topics. After automatically parsing the articles, we split them into paragraphs and eliminate ill-formed paragraphs matching any of the following conditions: (i) containing any special character repeated
more than three times (e.g., \%, *, etc.), (ii) not Arabic as classified by langdetect,~\footnote{\url{https://pypi.org/project/langdetect/}} and (iii) containing HTML tags. The paragraphs were de-duplicated using Cosine similarity, with a similarity >= 0.75 indicating duplication. 





 
The resulting news paragraphs were then manually annotated using 23 propaganda techniques, adopted from an existing taxonomy~\cite{piskorski2023news}. The annotation process consisted of two phases: \emph{(i)} in phase 1, three \textbf{annotators} individually annotated each paragraph, and \emph{(ii)} in phase 2, two expert annotators revised and finalized the annotations. Each annotator in this phase is referred to as a \textbf{consolidator}. To facilitate the annotation process, a platform was developed and a comprehensive annotation guideline in the native language (Arabic) was provided to annotators. Additionally, several training iterations were conducted before beginning the annotation task.


The annotation agreement for span-level annotation is $\gamma$ = 0.546. This $\gamma$ agreement metric is specifically designed for span/segment-level annotation tasks, taking into account the span boundaries (i.e., start and end) and their labels \cite{Mathet2015, mathet-2017-agreement}. Table~\ref{tab:data_split_span} reports the distribution of the span-level labels across the three dataset splits.

\begin{table}[t]
\centering
\resizebox{\columnwidth}{!}{%
\begin{tabular}{@{}lrrr@{}}
\toprule
\textbf{Technique} & \multicolumn{1}{l}{\textbf{Train}} & \multicolumn{1}{l}{\textbf{Dev}} & \multicolumn{1}{l}{\textbf{Test}} \\ \midrule
Appeal to Authority & 192 & 22 & 42 \\
Appeal to Fear-Prejudice & 93 & 11 & 21 \\
Appeal to Hypocrisy & 82 & 9 & 17 \\
Appeal to Popularity & 44 & 4 & 8 \\
Appeal to Time & 52 & 6 & 12 \\
Appeal to Values & 38 & 5 & 9 \\
Causal Oversimplification & 289 & 33 & 67 \\
Consequential Oversimplification & 81 & 10 & 19 \\
Conversation Killer & 53 & 6 & 13 \\
Doubt & 227 & 27 & 49 \\
Exaggeration-Minimisation & 967 & 113 & 210 \\
False Dilemma/No Choice & 60 & 6 & 13 \\
Flag Waving & 174 & 22 & 41 \\
Guilt by Association & 22 & 2 & 5 \\
Loaded Language & 7,862 & 856 & 1670 \\
Name Calling-Labeling & 1,526 & 158 & 328 \\
Obfuscation-Vagueness-Confusion & 562 & 62 & 132 \\
Questioning the Reputation & 587 & 58 & 131 \\
Red Herring & 38 & 4 & 8 \\
Repetition & 123 & 13 & 30 \\
Slogans & 101 & 19 & 24 \\
Straw man & 19 & 2 & 4 \\
Whataboutism & 20 & 4 & 4 \\ \midrule
\textbf{Total} & \textbf{13,212} & \textbf{1,452} & \textbf{2,857} \\ \bottomrule
\end{tabular}%
}
\caption{Distribution of the techniques in different data splits at the span level.}
\label{tab:data_split_span}
\end{table}
\section{Propagandistic Spans Annotation}
\label{sec:annot_exp}

\begin{figure*}[h]
    \centering
    \includegraphics[width=0.95\linewidth]{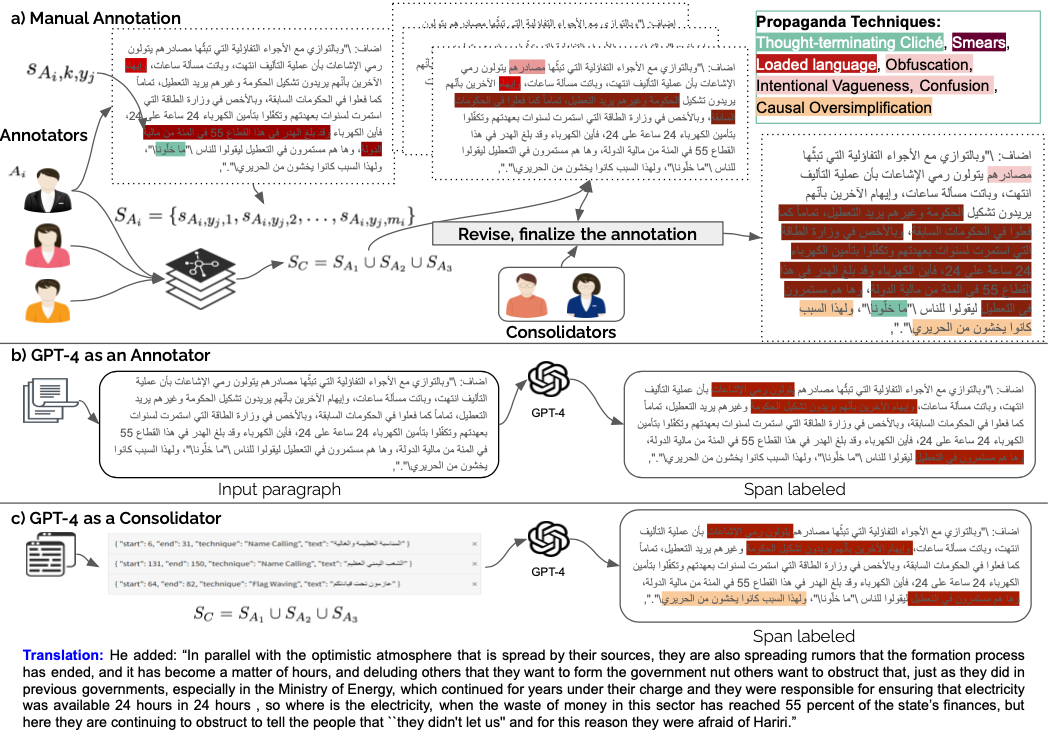}
    \caption{Existing span-level annotation process requiring human annotators and expert consolidators, while our proposed solution uses GPT-4 to support annotation and consolidation. 
    }
    \label{fig:pipeline}
\end{figure*}

In this section, we describe our annotation framework including the manual annotation steps used for dataset construction, and the use of GPT-4 for different annotation roles. Figure \ref{fig:pipeline} illustrates this framework. This section also describes a third annotation approach using fine-tuned models. 


\subsection{Manual Annotation}
The manual annotation process went through in two phases.  
For a given text \(\mathbf{x}=\{x_1, x_2, \ldots, x_n\}\) and a label (propaganda techniques) space \(\mathcal{Y}=\{y_1, y_2, \ldots, y_o\}\), each annotator $A_i$ provides a set of spans $S_{A_i}$ and each span is represented as $s_{A_i,y_j,k}$ , where $ k $ is the index of the span for the $ i $-th annotator and $y_j$ is the label. Note that $ k $ can range from 1 to the total number of spans identified by annotator $A_i$, and this total can be different for each annotator. Given this representation, for the $i^{th}$ annotator the set of spans is defined as
$S_{A_i} = \{ s_{A_i,y_j,1}, s_{A_i,y_j,2}, \ldots, s_{A_i,y_j,m_i} \}$
where $m_i$ is the total number of spans identified by annotator $A_i$ and \( y_j \) represents any label from the label space, where \( j \) can vary from 1 to \( o \). We combine the spans of all annotators into list 
$\mathcal{S_C}$ 
  that goes through the consolidation phase to finalize the annotations by consolidators.

To denote the labels (techniques) in a paragraph (input text $\mathbf{x}$) annotated by an annotator $A_i$, we define the following formulation:
$\mathbb{Y}_{A_i} = \bigcup_{j=1}^{p} A_{i,y_j}$
where \( \mathbb{Y}_{A_i} \) represents the set of all labels \(\{ y_1, y_2, \ldots, y_p \}\) annotated by \( A_i \), where \( p \) is the total number labels.
$\mathbf{Y}$ represents the list of labels from all annotators for a paragraph. 

\subsection{Annotation with GPT-4}

To formally define the problem, let us consider the model \(\mathcal{M}\), text input \(\mathbf{x}=\{x_1, x_2, \ldots, x_n\}\), and label space \(\mathcal{Y}\). The task of \(\mathcal{M}\) is to identify the text span \(\mathcal{S} = \{s_{1}, s_{2}, \ldots, s_{m_i}\}\) and an associated label for each span $s_i$, where $s_{i}=y \in \mathcal{Y}$.
The model is conditioned using instruction \( \mathcal{I} \), which describes both the task and the label space \( \mathcal{Y} \). This conditioning can occur in two scenarios: with a few-shot approach, utilizing labeled examples \((\mathbf{x}, \mathbf{y}) \in \mathcal{D}_l\), or in a zero-shot context, where labeled examples are not provided. $\mathcal{D}_l$ represents the labeled dataset. We formulated three levels of difficulty for the propaganda span annotation task using GPT-4. 

\begin{itemize}[leftmargin=*, nosep, topsep=0pt, after=\vspace{0pt}]
    \item \textbf{Instruction only (Annotator)}: 
    In this setup, the model is only provided with an instruction \( \mathcal{I} \) asking it to annotate the text \(\mathbf{x}\) by identifying the propaganda techniques used in it, and then extracting the corresponding spans \(\mathcal{S}\).
    \item \textbf{Span extractor (Selector)}: We offer additional 
    information for annotation and frame it as a
    span extraction problem. 
    The model is asked to select the techniques manifesting in text from the list $\mathbf{Y}$, and extract the matching text spans. 
    \item \textbf{Annotation consolidator (Consolidator)}: This setup is the most resource rich, where the model is asked to act as a consolidator, given 
    list $\mathcal{S_C}$ as provided by annotators. 
\end{itemize}

\subsection{Annotation with PLMs}
\label{ssec:annot_plm}
As a third annotation approach, we aim to train specialized models for the task, using manual and GPT-4 annotations to train a pre-trained language model (PLM). Fine-tuning PLMs, especially those following BERT~\cite{devlin2018bert} architecture, has dominated recent approaches for propaganda span detection~\cite{piskorski2023semeval,araieval:arabicnlp2024-overview}. We model our propaganda span detection and classification task as a span categorization problem, extended from typical token classification tasks like Named Entity Recognition. In this task, multiple labels can be assigned per token, as multiple propaganda techniques can appear as part of the same text span. Formally, we define the task as follows. Given an input token sequence \(\mathbf{x}=\{x_1, x_2, \ldots, x_n\}\) of length $n$, and a label (propaganda techniques) space \(\mathcal{Y}=\{y_1, y_2, \ldots, y_o\}\), the task is to predict $Y'$ of length 23 for each token, with one element for each label. An element $y'_i$ in $\mathcal{Y'}$, is either 0 or 1, indicating whether the token belongs to technique $y'_i$. 

For the model architecture, we select a BERT-based model, and apply a Sigmoid activation function at the output layer of the model, using a binary cross-entropy loss function. To decide whether a token $x_i$ belongs to category $y'_i$, we set a threshold $l$, and a model logit $>l$ indicates $x_i$ belongs to $y'_i$.
\section{Experimental Setup}
\label{sec:exp_setup}
 In this section, we describe the setup of the experiments and the evaluation approach followed to investigate the effectiveness of GPT-4 in playing different roles in the annotation process.   

\subsection{Datasets}
\label{sec:datasets}
\noindent 
\paragraph{Training and analysis:} For the main experiments in this study, we used the training subset of ArPro, ArPro$_{train}$, (discussed in Section~\ref{sec:dataset}) including 6,002 annotated paragraphs. In particular, we consider \textit{the annotations resulting from the consolidation phase as our gold standard labels in all experiments}.

\noindent 
\paragraph{PLM models training and testing:} We train four specialized models, one over each of the following training sets: ArPro$_{train}$, and GPT-4 predicted labels when acting as a consolidator, selector, and an annotator. We evaluate the trained models over two test sets: (i) the test subset of ArPro, ArPro$_{test}$, and (ii) a recent testing subset released with Task 1 of the ArAIEval shared task at the ArabicNLP 2024 conference~\cite{araieval:arabicnlp2024-overview}. The ArAIEval test subset includes both news paragraphs and tweets, and labeled following the same taxonomy of 23 propaganda techniques we adopt in this work. We chose to test against a second subset, to explore the models robustness and to put the performance of the specialized models, trained over GPT-4 predicted labels, in-context of relevant baseline systems from the shared task.

\subsection{Models} 
\noindent 
\paragraph{LLMs:} Across our different experiments, we used zero-shot learning using GPT-4 (32K, version gpt-4-0314, temperature=0) \cite{openai2023gpt4}. We chose this LLM due to its accessibility and superior performance compared to other open and closed  models~\cite{ahuja2023mega,abdelali2023benchmarking}.\footnote{Our initial experiments with another powerful closed model, Claude 3.5 Sonnet, showed that it performs similarly to GPT-4, so we opt to continue with GPT-4.}
\noindent 
\paragraph{PLM:} In our experiments in building specialized models for the task, we fine-tune AraBERTv0.2-large~\cite{antoun2020arabert},\footnote{\url{https://huggingface.co/aubmindlab/bert-large-arabertv02}} which is the most effective Arabic PLM to date over a variety of Arabic NLP tasks~\cite{antoun-etal-2021-araelectra}.\footnote{We have run the same set of experiments with another widely-used Arabic BERT model~\cite{safaya-etal-2020-kuisail} and observed similar patterns, thus, we only report results using AraBERT in this paper.}

\begin{table*}[t]
    \centering
    \resizebox{0.95\textwidth}{!}{%
    \begin{tabular}{l|p{18cm}}
     \toprule
     \textbf{Setup} & \textbf{Prompt} \\
     \hline
     \textbf{Annotator} & \textbf{Instruction (\( \mathcal{I} \)):} Label the "Paragraph" by the following propaganda techniques: [techniques list]. Answer exactly and only by returning a list of the matching labels from the aforementioned techniques and specify the start position and end position of the text span matching each technique. Use this template \{``technique'': , ``text'': , ``start'': , ``end'': \}

     \textbf{Paragraph}: \{\dots\}
     
     \textbf{Response}: 
    \\
    \hline
     \textbf{Selector} & \textbf{Instruction (\( \mathcal{I} \)):} Given the following ``Paragraph'' and ``Annotations'' showing propaganda techniques potentially in it. Choose the techniques you are most confident appeared in Paragraph from all Annotations and return a Response. Answer exactly and only by returning a list of the matching labels and specify the start position and end position of the text span matching each technique. Use this template Use this template \{``technique'': , ``text'': , ``start'': , ``end'': \}

     \textbf{Paragraph}: \{\dots\}
     
     \textbf{Annotations}: \(\mathbf{Y}\)

     \textbf{Response}: 
    \\
    \hline
     \textbf{Consolidator} & \textbf{Instruction (\( \mathcal{I} \)):} Given the following ``Paragraph'' and ``Annotations'' showing propaganda techniques potentially in it, and excerpt from the Paragraph where a technique is found. Choose the techniques you are most confident appeared in Paragraph from all Annotations and return a Response. Answer exactly and only by returning a list of the matching annotations.

     \textbf{Paragraph}: \{\dots\}
     
     \textbf{Annotations}: \(\mathcal{S_C}\)

     \textbf{Response}: 
    \\
    \bottomrule
    \end{tabular}%
    }
    \caption{Different prompts used to instruct GPT-4 to annotate input paragraphs by propaganda techniques and spans.}
    \label{tab:prompts}
\end{table*}

\subsection{Instruction}
Table~\ref{tab:prompts} lists the exact prompts used to invoke GPT-4 to act in its three different roles of interest in this work. During some pilot studies over the development subset, we have experimented with a variety of prompts for each of the roles before identifying the prompts we eventually used as they had the best performance. We also note that model generally performed really well in responding with the required JSON format of output. 


\subsection{PLM Fine-tuning} 
For \textit{each} of the training sets (listed in Section~\ref{sec:datasets}), we fine-tune the PLM for $ep$ epochs, setting the maximum sequence length to 256, a weight decay of 0.001, a train batch size of 16 and a learning rate of $1e-5$. \textit{For each of the four models we train}, the number of epochs $ep$ and the prediction threshold $l$ (Section~\ref{ssec:annot_plm}) are hyperparameters we tune over the development subset of ArPro, and report performance of the best model over the testing subset. For hyperparameter tuning, we follow a grid search approach, experimenting with $0.05<=l<=0.5$ (step=0.05) and  $5<=ep<=30$ (step=5).

\subsection{Evaluation}
\label{ssec:eval_measure}
We take two approaches to evaluate the performance of models for our tasks.

\paragraph{Standard System Evaluation.}
For both GPT-4 and fine-tuned models, we computed a modified version of the F$_1$ measure (macro- and micro-averaged) that accounts for partial matching between the spans across the gold labels and the predictions~\cite{propaganda-detection:WANLP2022-overview}. 


\paragraph{Inter-rater Agreement.}
As we are investigating GPT-4's ability as an annotator, we can also evaluate its performance through the computation of inter-rater agreement between its annotations and the gold labels from the dataset. 
We specifically computed \(\gamma\) \cite{Mathet2015, mathet-2017-agreement}, a measure used in similar tasks \cite{EMNLP19DaSanMartino}, which is designed for span/segment-level annotation tasks.

%

\section{Results and Discussion}
\label{sec:results}
To address our research questions, we ran each of the annotation setup prompts (Table~\ref{tab:prompts}) over all 6,002 paragraphs in the training split. Table~\ref{tab:all_results} shows the results of evaluating the post-processed model's outputs.

\begin{table}[h]
\centering
\setlength{\tabcolsep}{3pt}
\resizebox{0.8\columnwidth}{!}{%
\begin{tabular}{@{}lccc@{}}
\toprule
\textbf{Role} &\textbf{Micro-F$_1$} & \textbf{Macro-F$_1$} & \textbf{Span ($\gamma$)} \\ \midrule
Annotator & 0.050 & 0.045 & 0.247 \\
Selector & 0.137 & 0.144 & 0.477 \\
Consolidator & 0.671 & 0.570 & 0.609 \\ \bottomrule
\end{tabular}
}
\caption{Performance of GPT-4 (with its different roles) in propaganda span annotation using standard evaluation measures and annotation agreement.}
\label{tab:all_results}
\end{table}

As shown in Table~\ref{tab:all_results}, the more information provided to GPT-4 during annotation, the more improvement we observed in its performance.
In an information rich setup with GPT-4 as a ``consolidator'', where we used all the span-level annotations from three annotators, it led to significantly strong model performance. However, it should be noted that the task of a consolidator is not limited to deciding which of the initial annotations are the most accurate. They also had the freedom to modify the annotations by updating the annotation span length or by changing the label for a given span. 
As for annotation agreement, we can also see that the agreement scores were higher, when more information was provided to GPT-4 in the consolidator role, than the setups with less information. 



\begin{figure*}[t]
    \centering
    \includegraphics[width=0.85\linewidth]{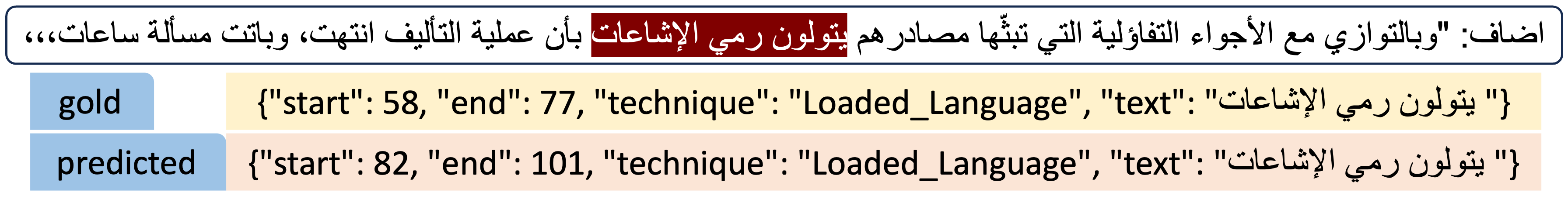}
    \caption{Example of wrongly generated span indices by GPT-4.
    }
    \label{fig:exm_index}
\end{figure*}

\paragraph{Incorrect start and end indices.}
In addition to detecting propaganda techniques, the model was required to provide the text spans matching these techniques (in the ``annotator'' and ``selector'' roles). Since a span might occur multiple times in a paragraph, with different context and propagandistic technique, 
the model should also specify the start and end indices of these spans
. We observed that although GPT-4 can correctly provide labels and extract associated text spans, it frequently generated indices not matching the corresponding spans in a paragraph. This lead to mismatch between the start and end indices of spans as compared to gold labels (As Figure~\ref{fig:exm_index} shows). 

\begin{table}[]
\centering
\resizebox{0.8\columnwidth}{!}{%
\begin{tabular}{lcc}
\toprule
\textbf{Role} &\textbf{Micro-F$_1$$_{orig}$} & \textbf{Micro-F$_1$$_{correct}$} \\
\hline
Annotator  & 0.050 &  0.117 \\
Selector & 0.137 &  0.297 \\
Consolidator & 0.671 & 0.670 \\ 
\bottomrule
\end{tabular}%
}
\caption{Performance of GPT-4 with ($_{correct}$) and without ($_{orig}$) span indices correction.}
\label{tab:results_standard_correct}
\end{table}
To overcome this problem, we a apply a post-processing step by assigning for each predicted span, the start and end indices of its first occurrence in a paragraph. Table~\ref{tab:results_standard_correct} reports the performance of GPT-4 following  this correction.
It reveals the severity of inaccurate span positions prediction. 
With the first two roles of the model, we observe the performance increasing by a factor of two with the applied correction. Interestingly, in its third role, as a consolidator, this problem did not manifest, as the model was only selecting annotations, including span and indices, from the list \(\mathcal{S_C}\) of all annotations. 

\begin{figure}[t]
    \centering
    \includegraphics[width=0.85\columnwidth]{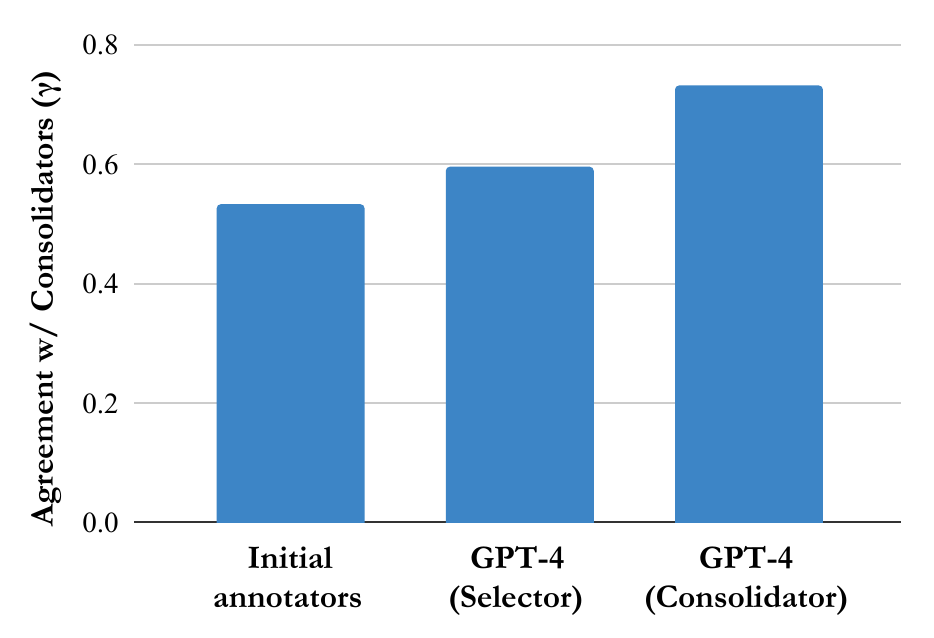}
    \caption{Agreement between consolidators and different types of annotators.}
    \label{fig:cons_agmnt}
\end{figure}

\paragraph{Agreement with consolidators.} We delve deeper into the quality of the model's annotations by comparing two values: (a) the agreement of the initial, less-experienced, annotators with the consolidators and (b) GPT-4 agreement with the consolidators (\textit{after start indices correction}). 
As Figure~\ref{fig:cons_agmnt} shows, GPT-4 has notably higher agreement with consolidators compared to initial annotators. It demonstrates a 38\% improved agreement when playing the role of a consolidator. \textit{These values demonstrate that GPT-4 achieves comparable or better agreement with the expert consolidators as compared to less experienced human annotators. Moreover, it shows that the model is learning from the given initial annotations to produce improved annotations, closer to the consolidators' performance.}


\begin{table}[h]
\centering
\resizebox{0.95\columnwidth}{!}{%
\begin{tabular}{lrl}
\hline
\textbf{Technique} & \multicolumn{1}{r}{\textbf{Annotator}}    &  \\ \hline
Causal Oversimplification         & 0.889 &  \\
Consequential Oversimplification  & 0.835 &  \\
\underline{Doubt}        & 0.815 &  \\
Obfuscation /Vagueness /Confusion & 0.791 &  \\
\underline{Appeal to Hypocrisy}         & 0.746 &  \\ \hline
 & \multicolumn{1}{r}{\textbf{Selector}}     &  \\ \hline
\underline{Doubt}        & 0.802 &  \\
Flag Waving        & 0.705 &  \\
\underline{Appeal to Hypocrisy}         & 0.660 &  \\
\underline{Loaded Language} & 0.654 &  \\
Slogans & 0.642 &  \\ \hline
 & \multicolumn{1}{r}{\textbf{Consolidator}} &  \\ \hline
False Dilemma /No Choice & 0.872 &  \\
\underline{Loaded Language} & 0.774 &  \\
Straw Man & 0.697 &  \\
\underline{Doubt}        & 0.695 &  \\
Name Calling /Labeling & 0.680 &  \\ \hline
\end{tabular}%
}
\caption{Agreement level (measured by $\gamma$) between GPT-4 and gold labels for top five techniques per role, with ($_{correct}$) span indices correction. Underlined are techniques appearing in at least two annotation roles.}
\label{tab:results_per_tec}
\end{table}

\paragraph{Per technique performance.}
Our next research question is: which propaganda techniques can GPT-4 annotate best? We looked at the top five per-technique agreement levels ($\gamma$) of the model's labels versus gold labels (Table~\ref{tab:results_per_tec}). Over all its roles, the model showed high agreement with expert annotators (consolidators) for three techniques: Doubt, Appeal to Hypocrisy and Loaded Language. It is interesting to see that GPT-4 was highly effective in annotation of the ``Doubt'' technique, which contradicts with a recent ranking of annotation difficulty of the same taxonomy, derived from humans' performance, in the same task across a multilingual dataset~\cite{stefanovitch2023holistic}. However, its strong performance with the other two techniques is inline with the aforementioned ranking. The model's ability to annotate ``Loaded Language'' is particularly useful, as it is the most prevalent technique in the dataset, appearing 7.9K times in the training split under investigation. Replacing human consolidators by GPT-4 to annotate for that technique can save tremendous time and cost. We believe these agreement levels give further evidence of the strong potential of employing GPT-4 as a propaganda span annotator, at least for some techniques. This analysis also provides data needed to inform decisions on which stages of annotation we can inject LLMs like GPT-4.

\paragraph{Performance of the specialized model.}
To gain a deeper understanding of the effect of using GPT-4 as an annotator, we use the labels provided by the model in its different annotation roles to train specialized models for the task. 

Table~\ref{tab:plm_results} compares the performance of AraBERT on the ArPro test subset, when fine-tuned with different training sets. We also compare its performance to two baselines: (i) a random baseline, that randomly assigns propaganda techniques to random spans of text in a paragraph~\cite{propaganda-detection:WANLP2022-overview}, and (ii) prompting GPT-4 to predict labels on the test set using the first prompt in Table~\ref{tab:prompts}.

\begin{table}[]
\centering
\resizebox{\columnwidth}{!}{%
\begin{tabular}{lllll}
\toprule
\textbf{Model}  & \textbf{Train Set}     & \textbf{$ep$} & \textbf{$l$} & \textbf{Micro-F$_1$}          \\
\hline
Random & - & -  & -  & 0.010 \\
GPT-4 & - & -  & -  & 0.117   \\
AraBERT & GPT-4$_{Annotator}$    &   20 &   0.10   &    0.127\\    
AraBERT & GPT-4$_{Selector}$     & 25 & 0.30 & 0.236  \\     
AraBERT & GPT-4$_{Consolidator}$ & 25 & 0.15 & 0.335  \\
AraBERT & ArPro$_{train}$      & 30 & 0.25 & 0.387  \\
\bottomrule
\end{tabular}%
}
\caption{Performance of the PLM when fine-tuned on different training sets, and tested on ArPro$_{test}$. Span indices correction was applied to all GPT-4 predictions. $ep$: number of training epochs, $l$: prediction threshold.}
\label{tab:plm_results}
\end{table}


Results in Table~\ref{tab:plm_results} lead to several conclusions. First, models fine-tuned for the task in all four setups outperform GPT-4 when directly used to detect and label propagandistic spans (2$^{nd}$ row). This motivates the need for specialized models for such complex span categorization task. Second, compared to training the model on the gold labels (6$^{th}$ row), training the model on GPT-4's labels when serving as a consolidator (5$^{th}$ row) reduces performance by only 13\%, further supporting our conclusions on the value of using GPT-4 as a consolidator.  

We further evaluate the quality of GPT-4 annotations for model fine-tuning, by testing the trained models over a second testing subset, ArAIEval24T1$_{test}$~\cite{araieval:arabicnlp2024-overview}. We compare the models performance to the top performing system from the shared task, CUET\_sstm~\cite{araieval-arabicnlp:2024:task:CUET_sstm}.

Results in Table~\ref{tab:plm_results_araieval} endorse using GPT-4$_{Consolidator}$ labels to train specialized models, as it lead to relative improvement over the baseline by 11\%. Furthermore, we observe a 35\% relative improvement over the top team from the shared task, when we train our model on the ArPro$_{train}$ subset; achieving state-of-the-art for the propaganda span detection and categorization task over this large-scale Arabic testing dataset.

\begin{table}[]
\centering
\resizebox{0.8\columnwidth}{!}{%
\begin{tabular}{lll}
\toprule
\textbf{Model}  & \textbf{Train Set}  & \textbf{Micro-F$_1$}  \\
\hline
CUET\_sstm   & -  & 0.300 \\
AraBERT & GPT-4$_{Annotator}$    & 0.124 \\
AraBERT & GPT-4$_{Selector}$     & 0.257 \\
AraBERT & GPT-4$_{Consolidator}$ & 0.334 \\
AraBERT & ArPro$_{train}$      & 0.406 \\
\bottomrule
\end{tabular}%
}
\caption{Performance of the fine-tuned PLM when tested on ArAIEval24T1$_{test}$.}
\label{tab:plm_results_araieval}
\end{table}
\section{Conclusions}
\label{sec:conclusions}
In this study, we first investigate GPT-4's ability to play different roles in detecting propagandistic spans and annotating them in Arabic news paragraphs. We investigate if GPT-4 can be used as an annotator when provided with sets of information of varied richness, which represents an increased cost and effort in hiring human annotators. Moreover, we study the value of GPT-4's labels when used to train specialized models for the task. Our experimental results suggest that providing more information significantly improves the model's annotation performance and agreement with human expert consolidators. The study also reveals the great potential of the model to replace consolidators, for some propaganda techniques. Finally, we find that we can train effective models using labels provided by GPT-4 when acting as a consolidator. We offer an in-depth analysis of the model's performance across various annotation stages, facilitating a more informed adoption of this annotation approach. Future research will explore additional models and learning setups.

\section{Limitations}
\label{sec:limitations}
The current version of our work focuses on the analysis and evaluation of GPT-4 specifically limited to Arabic. For this study, we chose to use an Arabic dataset because annotated labels from multiple annotators are available, which are often difficult to obtain. We have evaluated only a closed LLM, as it is currently the most effective model for a large variety of NLP tasks and languages, as reported in a myriad of studies. Moreover, we have ran experiments with large and effective open models for the task, which revealed that they are either unable to understand the task or showed more inferior performance compared to the closed LLM. 

\section*{Ethics and Broader Impact}

\label{sec:ethics}
We do not foresee any ethical issues in this study. We utilized an in-house dataset consisting of paragraphs curated from various news articles. Our analysis will contribute to the future development of datasets and resources in a cost-effective manner. Human annotators identity will not be shared and cannot be inferred from the annotations we plan to release. We would like to warn users to carefully use the annotations that we plan to release. Its misuse (e.g., using them to generate similar content) may lead to potential risks. 

\section*{Acknowledgments}
The work of M. Hasanain and F. Ahmad is supported by the NPRP grant 14C-0916-210015 from the Qatar National Research Fund part of Qatar Research Development and Innovation Council (QRDI). The findings achieved herein are solely the responsibility of the authors.
\bibliography{bib/all,bib/arab_prop}

\end{document}